\documentclass{article}
\usepackage{spconf}
\usepackage{cite}
\usepackage{amsmath,amssymb,amsfonts}
\usepackage{graphicx}
\usepackage{textcomp}
\usepackage{lipsum, color}
\usepackage{graphics}
\usepackage{titlesec}
\usepackage{subfig}
\usepackage{graphicx}
\usepackage{xcolor}
\usepackage{titlesec}
\usepackage{subfig}
\usepackage{graphicx}
\usepackage{mathtools}
\usepackage{siunitx}
\usepackage{soul}
\usepackage{balance}
\usepackage{algpseudocode}
\usepackage{cite}
\usepackage[font=footnotesize]{caption}
\usepackage{tabularx}
\usepackage{array}
\usepackage{floatrow}
\usepackage[english]{babel}
\usepackage[utf8x]{inputenc}
\usepackage{amsmath}
\usepackage{xcolor}
\usepackage{tikz}
\usepackage{blindtext}
\usepackage{enumitem}
\usepackage[T1]{fontenc}
\usepackage{flushend}
\usepackage{multirow}
\usepackage{graphicx} % Needed for \rotatebox
\usepackage{rotating}
\usepackage{booktabs}
\def\BibTeX{{\rm B\kern-.05em{\sc i\kern-.025em b}\kern-.08em
    T\kern-.1667em\lower.7ex\hbox{E}\kern-.125emX}}

\begin{document}
\title{Conformalized Multimodal Uncertainty Regression and Reasoning}

\name{Domenico Parente, Nastaran Darabi, Alex C. Stutts, Theja Tulabandhula, and Amit Ranjan Trivedi}

\address{University of Illinois at Chicago (UIC), Chicago, USA}
\maketitle

\begin{abstract}
This paper introduces a lightweight uncertainty estimator capable of predicting multimodal (disjoint) uncertainty bounds by integrating conformal prediction with a deep-learning regressor. We specifically discuss its application for visual odometry (VO), where environmental features such as flying domain symmetries and sensor measurements under ambiguities and occlusion can result in multimodal uncertainties. Our simulation results show that uncertainty estimates in our framework adapt sample-wise against challenging operating conditions such as pronounced noise, limited training data, and limited parametric size of the prediction model. We also develop a reasoning framework that leverages these robust uncertainty estimates and incorporates optical flow-based reasoning to improve prediction prediction accuracy. Thus, by appropriately accounting for predictive uncertainties of data-driven learning and closing their estimation loop via rule-based reasoning, our methodology consistently surpasses conventional deep learning approaches on all these challenging scenarios--pronounced noise, limited training data, and limited model size--reducing the prediction error by 2--3$\times$. 
\end{abstract}
\begin{keywords}
Conformal inference. Visual odometry.
\end{keywords}

\section{Introduction}
Cutting-edge deep learning frameworks are striving not only for accurate point predictions but also to express predictive uncertainties. This is achieved by incorporating uncertainty-aware learning and prediction techniques such as Bayesian neural networks, Gaussian processes, Monte Carlo dropout, variational inference, \textit{etc.} \cite{goan2020bayesian,schulz2018tutorial,gal2016dropout,blei2017variational}. However, the robustness of learning uncertainty-aware prediction comes at a considerable computational cost. For example, methods like Monte Carlo dropout require multiple model samplings to capture the predictive model's distribution, where each sampling entails running model inference. For applications such as autonomous drones, where predictions must be made in real-time and with limited onboard resources, the computational challenges for uncertainty-aware predictions thus become excessive \cite{shukla2022mc,shukla2021ultralow}.

To address these limitations of computationally efficient uncertainty-aware deep learning, there has been a notable surge of interest in \textit{conformal inference} \cite{einbinder2022training,karimi2023quantifying,alaa2023conformalized,romano2019conformalized,stutts2023lightweight}. This approach offers a systematic framework for estimating predictive uncertainty intervals by incorporating calibration measures into training procedures. The process entails training a learning model on a labeled dataset and utilizing calibration data to construct a predictive uncertainty region around each test instance. In the case of classification problems, conformal inference adapts them into a minimal set prediction task, where there is a high confidence that the true class lies within the prediction set. Likewise, regression problems transform into minimal interval prediction tasks, ensuring that the true value falls within the interval with high confidence. Consequently, by employing conformal prediction, learning models can provide point predictions \textit{as well as} a reliable measure of confidence associated with each prediction. A key computational advantage of conformal inference is its explicit focus on the uncertainty bounds rather than computing the distribution, which enhances its computational efficiency.

\begin{figure}[t!]
    \centering
    \includegraphics[width=\linewidth]{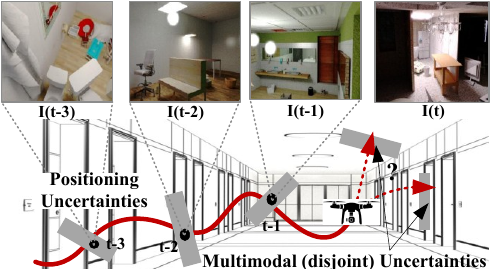}
    \caption{\textbf{Multimodal Uncertainties:} Environmental features such as flying domain symmetries and sensor measurements under ambiguities and occlusion can result in multimodal (disjoint) uncertainty bands in practical applications such as visual odometry. We develop lightweight multimodal uncertainty regression and reasoning to address these practical challenges.}
    \label{fig:Introduction}
\end{figure}

\begin{figure*}[t!]
    \centering
    \includegraphics[width=\linewidth]{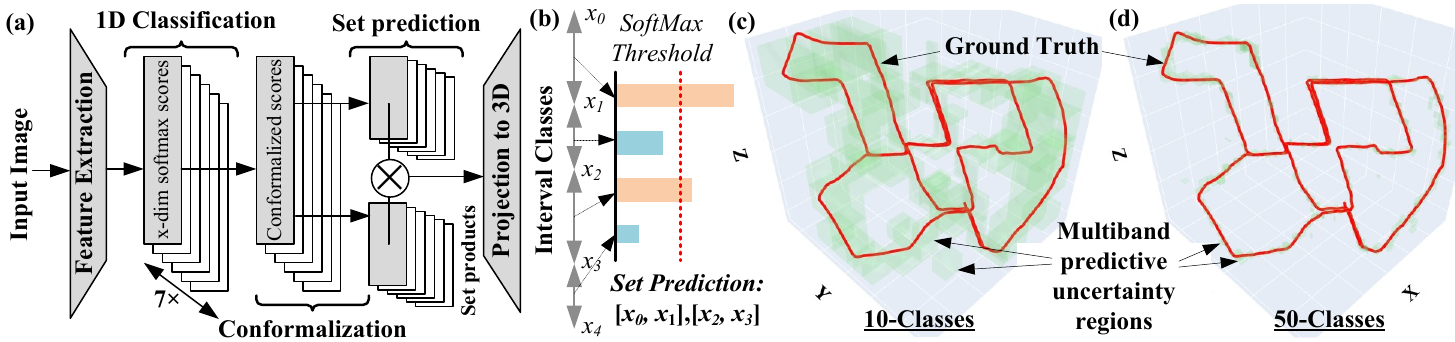}
    \caption{\textbf{Multimodal Conformalized Regression:} \textbf{(a)} Our framework for embedding conformalization in a deep learning regressor. \textbf{(b)} Example demonstration of our scheme by conformalization of softmax scores of multiple interval classes resulting in multimodal (disjoint) uncertainty interval prediction. Sample demonstration of our scheme for \textbf{(c)} 10-class and \textbf{(d)} 50-class multi-head configuration. }
    \label{fig:Introduction}
\end{figure*}

\textit{Yet}, a notable drawback of current conformal inference methods for regression is that they can only predict contiguous (single mode) uncertainty intervals \cite{romano2019conformalized,feldman2021improving,tang2022nonparametric,stankeviciute2021conformal}. Meanwhile, predictive uncertainties can be multi-modal in many cases. For example, in Fig. 1, consider an autonomous drone navigating a hallway with multiple similar rooms, doors, and intersections. Determining the vehicle's position in such surroundings will likely involve multimodal uncertainties due to environmental symmetries. Additionally, sensor measurements can be noisy or prone to occlusions, resulting in multiple plausible interpretations. Hence, computationally efficient multimodal uncertainty extraction is crucial. 

Addressing these limitations, our work makes the following key contributions: Using visual odometry (VO) as a driving application, we present a novel conformalized multimodal uncertainty regression by transforming regression into a set prediction problem. Results show that the uncertainty estimates adapt sample-wise against varying operating conditions such as input noise, limited training data, and the model's limited parametric size. We also present multimodal uncertainty reasoning that leverages the robust uncertainty estimates and closes the estimation loop via rule-based reasoning. Specifically, under challenging scenarios such as extreme noise, limited training data, and limited computational constraints, our integrated methodology consistently outperforms traditional deep learning by a 2--3$\times$ improvement in regression accuracy.

\section{Multimodal Uncertainty Regression}
Overcoming the limitations of current conformalized regression methods \cite{romano2019conformalized,feldman2021improving,tang2022nonparametric} and using VO as a driving application, we present a novel conformalized multimodal uncertainty regression. VO is a predominant computer vision technique in robotics to estimate the pose of a mounted camera on a drone/robot \cite{nister2004visual,scaramuzza2011visual,he2020review}. Deep learning-based VO employs neural networks to automatically learn and extract features from consecutive image frames to estimate the camera's pose. 

Our method reformulates the learning-based pose regression task as a \textit{set prediction problem}. In this context, the predicted set can represent uncertainty intervals that are not confined to being contiguous. To achieve this transformation, we extract a calibration set from the initial training data, enabling the conversion of the pose classification challenge into a set prediction task through conformalization, as depicted in Fig. 2(b). The first step involves segmenting the drone's navigational space into $K$ unique sets, spanning both position and orientation (or pose) dimensions. This segmentation is achieved using a non-uniform space discretization, which is informed by the training set trajectories within each dimension. These trajectories are split into $K$ quantiles to establish class boundaries. The quantile-based non-uniform discretization groups frequently visited poses into more confined spatial intervals. This allows for their identification with enhanced precision. The outcome of this space encoding process is a one-hot encoded matrix, contingent on the $K$ classes. In the subsequent stage, a neural network extracts features from input images. These features are then relayed to a multi-head classifier, as illustrated in Fig. 2(a). Each classifier head generates the softmax scores for interval classes along its respective pose dimension.

Based on the softmax scores, our procedure then targets the \textit{sample-adaptive minimum prediction set} $C(X) \subset \{1,...,K\}$ such that the correct class resides within the set with $1 - \alpha$ probability, where $\alpha$ denotes an arbitrary miscoverage rate (e.g., 10\%). This marginal coverage can be expressed as $1 - \alpha \leq \mathbb{P}\{Y \in C(X) \mid X=x\} \leq 1 - \alpha + \frac{1}{n+1}$, where $n$ denotes the size of the calibration set. Conformal scores arise by deducting the softmax output of the appropriate class for every input from one.
Finally, $\hat{q}$ is computed as the $\left \lceil{\frac{(n+1)(1-\alpha)}{n}}\right \rceil$ empirical quantile of these conformal scores, and the conformalized prediction set $C(X)$ is formed by incorporating classes with softmax scores greater than $(1-\hat{q})$. The product of the predicted set of classes along each pose dimension produces the net uncertainty regions, which are not necessarily contiguous. 

\begin{figure}[t!]
    \centering
    \includegraphics[width=\linewidth]{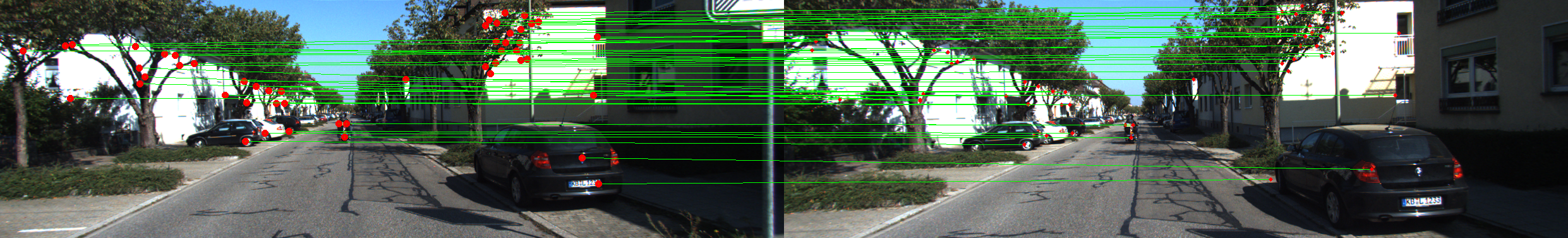}
    \caption{Optical flow-based correspondence between consecutive frames.}
    \label{fig:Introduction}
\end{figure}

Figs. 2(c-d) shows the example multimodal uncertainty regions predicted using the above procedure on the KITTI dataset \cite{Geiger2013IJRR}. The uncertainty regions are cuboid-shaped due to the product of intervals along respective pose dimensions. In Fig. 2(d), the uncertainty cuboids contract as the number of classes $K$ increases from ten to fifty due to higher precision interval classification. The multimodality and non-contiguity of uncertainty intervals are evident in the figure. The uncertainty estimates also adapt to drone's motion intricacies. For example, the predictive uncertainty increases at sharp turns. 
%In our prior work \cite{stutts2023lightweight}, we introduced the above framework; however, the potential for multimodal uncertainty estimation was not elaborated. In this work, we extended the technique for multimodal uncertainty estimation \textit{as well as} reasoning, which results in excellent noise tolerance, sample efficiency, and parametric efficiency compared to classical VO, as will be discussed in Sec. 4. 

\begin{figure}[t!]
    \centering
    \includegraphics[width=\linewidth]{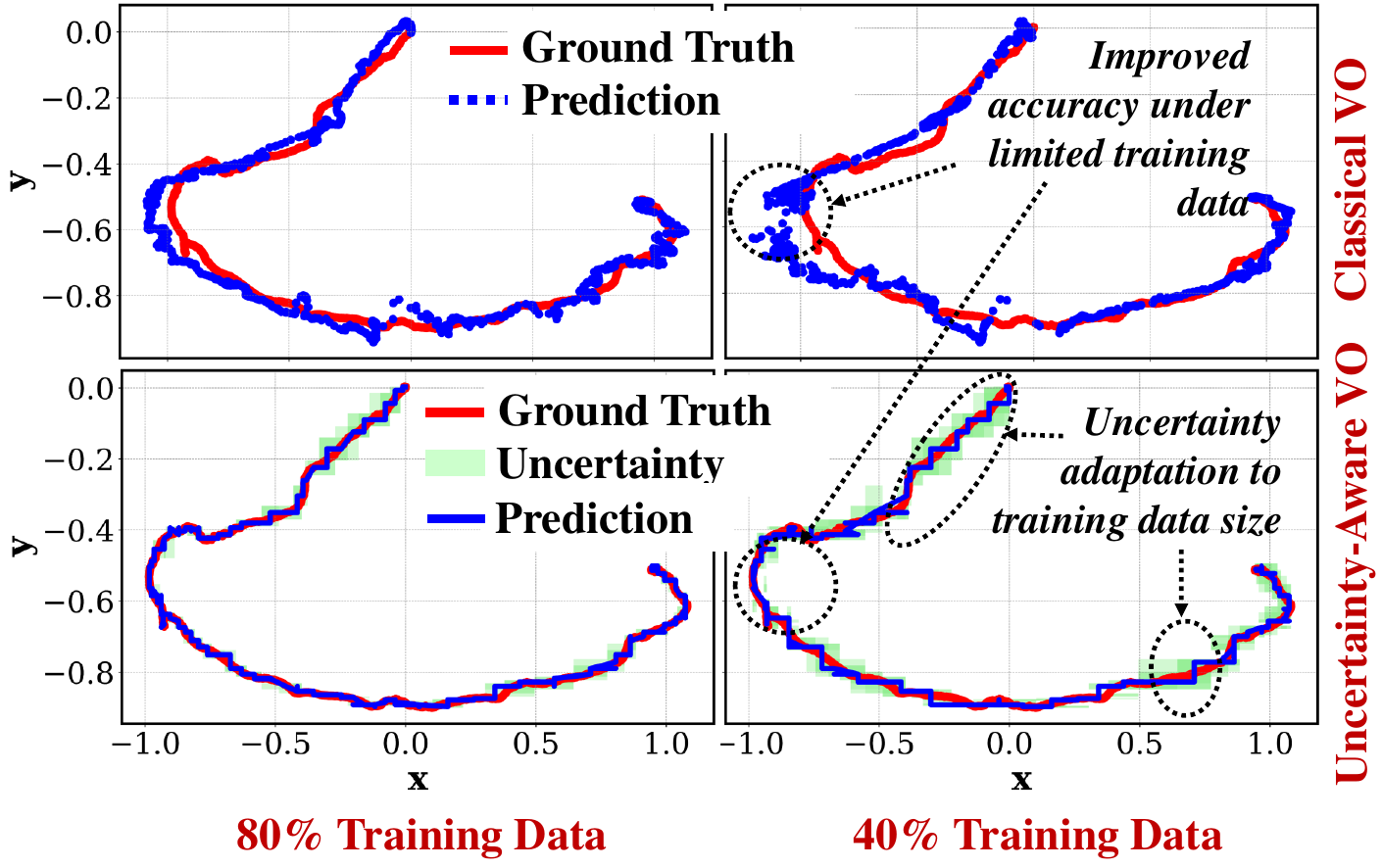}
    \caption{\textbf{Sample Efficiency:} In the right panel, the proposed framework maintains high accuracy despite a significant reduction in training data. Predictive uncertainties adaptively increase in our framework with lower training data. \textbf{\textit{[Testcase: RGB-D dataset \cite{RGBD}, first test sequence, 50-class multihead]}}}
\end{figure}

\section{Multimodal Uncertainty Reasoning}

\subsection{Relative Motion Estimation}
For reasoning among multimodal uncertainty estimates, our framework determines the relative motion between two frames using the Harris corner detector \cite{derpanis2004harris} as shown in Fig. 3. The process identifies key feature points, often located at angles or regions with intensity discontinuities in the first image. Subsequently, the Lucas-Kanade algorithm \cite{baker2004lucas} pinpoints the corresponding key points in the second frame. The algorithm presumes minimal movement of interest points between consecutive frames -- which is a reasonable assumption under a sufficiently high sampling rate and consistent brightness. Under this assumption, starting from the optical flow equation for each point $(x,y)$: $I(x + dx, y + dy, t + dt) \approx I(x, y, t)$, a linear approximation of the optical flux in a neighborhood of the points of interest is computed using a Taylor series. 

The algorithm assumes that the change in brightness of a pixel of the scene is totally compensated by the gradient of the scene itself, i.e., $I_x u + I_y v + I_t = 0$. Therefore, the displacement vector $[dx, dy]$ can be estimated by minimizing the squared error between points in the initial image and their counterparts in the subsequent frame. Based on the computed optical flow, points of interest are continuously updated, discarding those with significant tracking discrepancies, which also yields the key feature points for the second frame. 

Based on the feature correspondence, the relative motion between two poses can be extracted by exploiting the essential matrix $E$, given by $x_{1}^T \cdot E \cdot x_{0} = 0$, where $x_{0}$ is a corresponding point in the first image and $x_{1}$ is the corresponding point in the destination image. By factoring the matrix using singular value decomposition (SVD), rotation matrix $R$ and translation unit vector $t$ can be extracted as in \cite{hartley2007global}. Notably, essential matrix-based rotation and translation estimates often prove challenging in practice due to their sensitivity to small errors in point correspondences, leading to significant inaccuracies in the derived transformations. Despite these challenges, we leverage them solely for discerning uncertainty intervals, i.e., not as the main predictor. As elaborated in the subsequent section, such integrated prediction combining uncertainty-aware deep learning and optical flow-based reasoning yields impressive predictive accuracy across benchmarks, including noise robustness, sample efficiency, and parametric efficiency.    

\begin{figure}[t!]
    \centering
    \includegraphics[width=\linewidth]{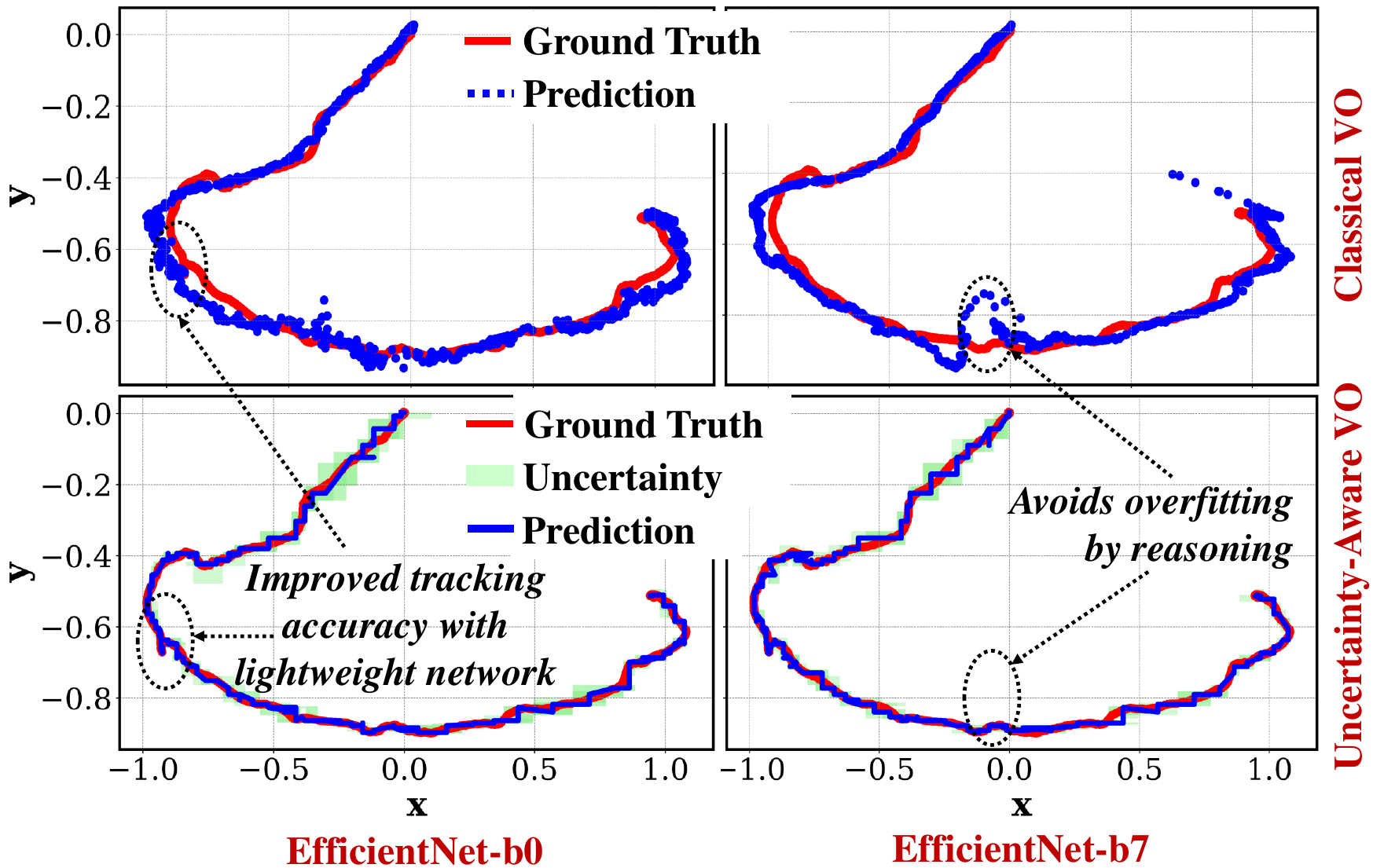}
    \caption{\textbf{Parametric Efficiency:} In the left panel, the proposed uncertainty-aware prediction framework maintains high accuracy despite a smaller network size of four million parameters. In the right panel, the proposed framework reduces overfitting risks by uncertainty-oriented predictions. \textbf{\textit{[Testcase: RGB-D dataset \cite{RGBD}, first test sequence, 50-class multihead]}}}
    \label{fig:Introduction}
\end{figure}

\vspace{-7pt}
\subsection{Uncertainty Discrimination}
After determining the relative rotation and translation between two frames, the next step involves discerning among multiple uncertainty intervals to ascertain the optimal value.
We initiate this by computing the mean value within each uncertainty interval to approximate the mean directions of displacements. Given that distinct predicted bounds exist for every dimension, we evaluate all possible combinations to identify the most fitting value. The algorithm takes the first pose with the highest softmax score from the prediction model and then iteratively searches for the best pose of the next frame. To find the best prediction of the orientation corresponding to the next frame, first, we calculate the corresponding rotation matrix as $R_{next}= R_{relative} \cdot R_{previous}$, then we transform it into a quaternion $q_{next}$ and at this point we look for the best prediction by solving:
\begin{align}
\underset{q_{\text{predicted}}\in Q}{\text{minimize}} \quad \lVert q_{\text{predicted}} - q_{\text{next}} \rVert
\end{align}
where $Q$ is the set of all quaternions describing the orientation of the camera that took the frame. Likewise, to find the best prediction of the spatial position corresponding to the next frame, we use the translation direction $t_{relative}$ calculated in the relative motion estimation phase and solve:
\begin{align}
\underset{t_{\text{predicted}}\in T}{\text{minimize}} \quad \left\lVert t_{\text{relative}} - \frac{t_{\text{predicted}} - t_{\text{previous}}}{\left| t_{\text{predicted}} - t_{\text{previous}} \right|} \right\rVert
\end{align}
where $T$ is the set of all translation vectors describing the position of the camera that took the frame. Solving these optimization steps selects the uncertainty interval with the highest likelihood for mean prediction and uncertainty discrimination on successive predictions.

\section{Simulation Results}
In Table 1 and Figs. 4-6, we compare the proposed framework to a conventional regression model on \textit{sample efficiency} (i.e., at varying sizes of labeled training data), \textit{parametric efficiency} (i.e., the necessary deep learning model size), and \textit{noise robustness}. For the comparative study, we only modify the training procedure from conventional to conformalized uncertainty-based reasoning while keeping all other settings, such as feature extraction, training data, \textit{etc.} the same. 

\begin{figure}[t!]
    \centering
    \includegraphics[width=\linewidth]{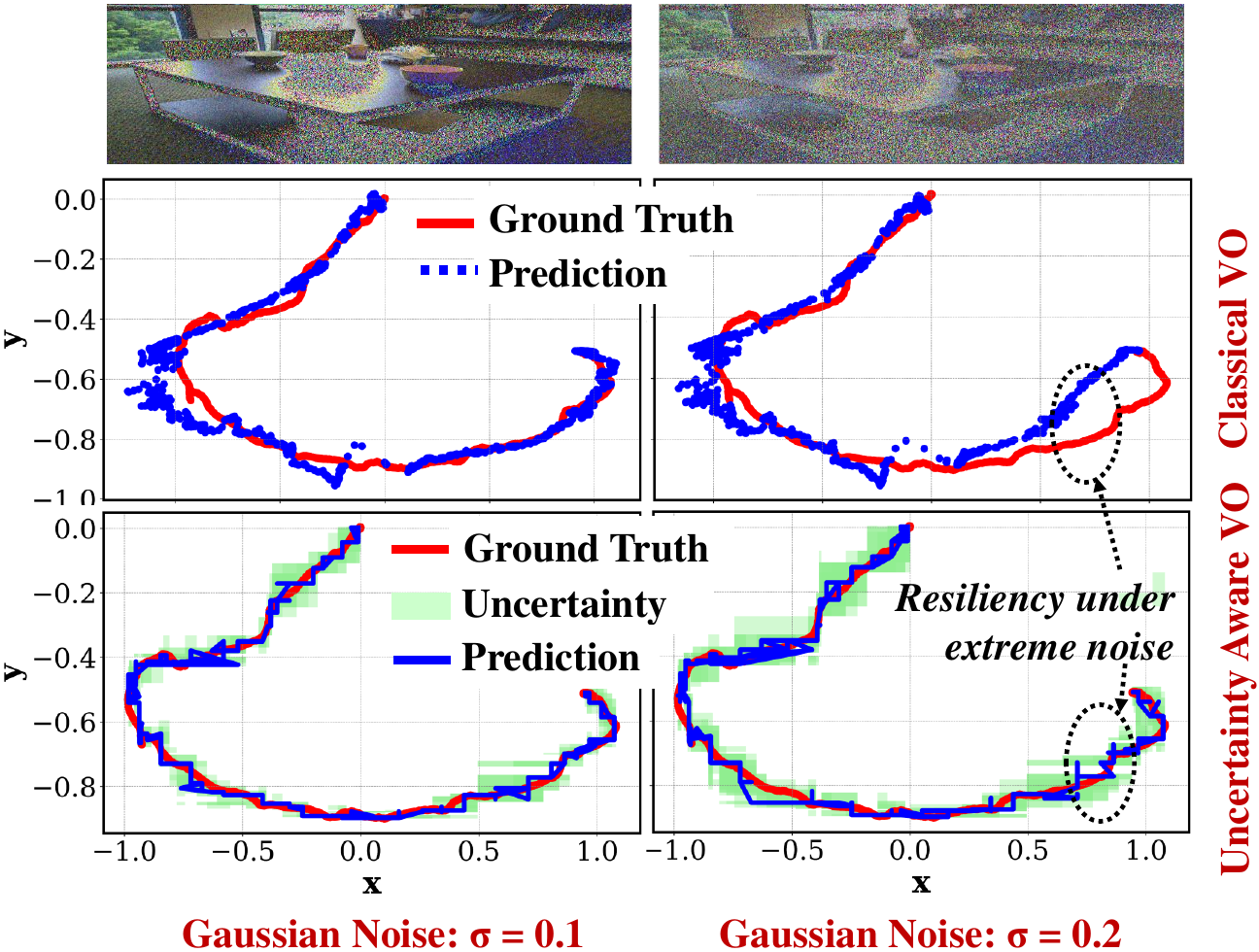}
    \caption{\textbf{Noise Robustness:} In the right panel, the proposed framework shows exquisite resilience to extreme noise. Predictive uncertainty levels appropriately adapt to input image noise levels. \textbf{\textit{[Testcase: RGB-D dataset \cite{RGBD}, first test sequence, 50-class multihead]}}}
\end{figure}

\vspace{3pt}
\noindent \textbf{Sample Efficiency:} Gathering sufficient labeled data for applications like VO poses significant challenges \cite{li2020self}. Under these practical challenges, Fig. 4 compares the training sample efficiency of the proposed framework against the conventional. In the right panel, the proposed framework maintains high accuracy even when the training data reduces to 40\% of the original. Notably, under limited training data, our framework appropriately expresses higher predictive uncertainties, as seen in left \textit{vs.} right panels. 

\begin{table}[t!]
    \footnotesize
    \centering
    \caption{Comparison of RMSE (Root Mean Square Error) of Pose Prediction Trajectories in Classical Inference vs. Our Uncertainty-Aware Inference.}
    \begin{tabularx}{\columnwidth}{Xccc}
        \toprule
        \textbf{Metric} & \textbf{Conformal} & \textbf{Classical} & \textbf{Improvement} \\
        \midrule
        40\% total data & 0.0534 & 0.136 & 2.5$\times$ \\
        80\% total data & 0.026 & 0.078 & 3$\times$ \\
        \cmidrule(lr){1-4}
        Efficientnet-b0 (4M) & 0.029 & 0.083 & 2.9$\times$ \\
        Efficientnet-b3 (10.7M) & 0.031 & 0.077 & 2.5$\times$ \\
        Efficientnet-b7 (63.8M) & 0.028 & 0.088 & 3.1$\times$ \\
        \cmidrule(lr){1-4}
        No noise & 0.050 & 0.136 & 2.7$\times$ \\
        Gaussian ($\sigma = 0.05$) & 0.055 & 0.126 & 2.3$\times$ \\
        Gaussian ($\sigma = 0.1$) & 0.073 & 0.132 & 1.8$\times$ \\
        Gaussian ($\sigma = 0.2$) & 0.099 & 0.201 & 2$\times$ \\
        \bottomrule
    \end{tabularx}
    \label{tab:your_label}
\end{table}

\vspace{3pt}
\noindent \textbf{Parametric Efficiency:} Smaller deep networks are essential for edge computing of VO to minimize computational and memory demands, enabling real-time processing on resource-constrained devices. In Fig. 5, we compare the performance of our framework against conventional deep learning regression at two variants of feature extractor--EfficientNet-b0 (with four million parameters) and EfficientNet-b7 (with $\sim$sixty-four million parameters) \cite{tan2019efficientnet}. Across these diverse configurations, the performance of the conformal models remained strikingly consistent despite the substantial variation in model parameters. While deep learning-based regressors underfit or overfit at too small or too large a model, in the proposed framework, the parametric resources are used under the guidance of uncertainty calibration. Our framework's mean prediction curves in the left and right panels are consistent. In contrast, the uncertainty bounds reduce with higher parameters in the right panel for EfficientNet-b7.

\vspace{3pt}
\noindent \textbf{Noise Robustness:} Input noise tolerance is crucial for VO to ensure accurate pose estimation under challenging environmental conditions such as low light scenarios. In Fig. 6, we compare predictions from both procedures while injecting Gaussian noise to input image pixels under varying noise variance ($\sigma$). Notably, the proposed framework also shows exquisite tolerance to input image noise and responds to higher noise levels in input images by expanding predictive uncertainty intervals.

Table 1 compares the RMSE (Root Mean Square Error) of pose prediction trajectories on the first testing sequence of RGB-D Scenes dataset \cite{RGBD} in classical deep learning-based regression against our procedure. By systematically acknowledging the variabilities in training data, model size, and input noise, our multimodal uncertainty-aware procedure improves prediction accuracy by $\sim$2-3$\times$ across all these diverse variations. When faced with multiple modes of uncertainty (i.e., disjoint uncertainty intervals), the reasoning framework is able to evaluate each mode's implications separately, thus capturing these complexities more accurately than an unimodal distribution, which otherwise oversimplifies.

\vspace{-10pt}
\section{Conclusions}
We presented a novel conformalized multimodal uncertainty regressor and reasoning framework for VO. The proposed framework demonstrated $\sim$2-3$\times$ lower prediction error than conventional deep learning under challenging scenarios such as lower training data, model size, and extreme input noise.

%\balance
\bibliographystyle{IEEEtran}
\bibliography{main.bib}
\end{document}